\documentclass[sigconf,screen]{acmart}
\usepackage{url}
\usepackage{xcolor}
\definecolor{newcolor}{rgb}{.8,.349,.1}
\usepackage{enumitem}
\usepackage{colortbl}
\usepackage{float}
\usepackage[ruled]{algorithm2e}
\usepackage{subcaption}
\usepackage{multirow}
\usepackage{enumitem}
\usepackage{listings}
\definecolor{mygray}{gray}{.9}
\definecolor{ForestGreen}{RGB}{34,139,34}

\usepackage{tcolorbox}
\copyrightyear{2024}
\acmYear{2024}
\setcopyright{rightsretained}
\acmConference[MM '24]{Proceedings of the 32nd ACM International Conference on Multimedia}{October 28-November 1, 2024}{Melbourne, VIC, Australia}
\acmBooktitle{Proceedings of the 32nd ACM International Conference on Multimedia (MM '24), October 28-November 1, 2024, Melbourne, VIC, Australia}
\acmDOI{10.1145/3664647.3689145}
\acmISBN{979-8-4007-0686-8/24/10}

\makeatletter
\gdef\@copyrightpermission{
  \begin{minipage}{0.3\columnwidth}
   \href{https://creativecommons.org/licenses/by/4.0/}{\includegraphics[width=0.90\textwidth]{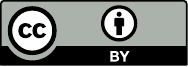}}
  \end{minipage}\hfill
  \begin{minipage}{0.7\columnwidth}
   \href{https://creativecommons.org/licenses/by/4.0/}{This work is licensed under a Creative Commons Attribution International 4.0 License.}
  \end{minipage}
  \vspace{5pt}
}
\makeatother

\newcommand\datasetabbr{AV-Deepfake1M}

\begin{document}

\title{1M-Deepfakes Detection Challenge}


\author{Zhixi Cai}
\orcid{0000-0001-7978-0860}
\affiliation{%
  \institution{Monash University}
  \city{Melbourne}
  \country{Australia}}
\email{zhixi.cai@monash.edu}

\author{Abhinav Dhall}
\orcid{0000-0002-2230-1440}
\affiliation{%
  \institution{Flinders University}
  \city{Adelaide}
  \country{Australia}}
\email{abhinav.dhall@flinders.edu.au}

\author{Shreya Ghosh}
\orcid{0000-0002-2639-8374}
\affiliation{%
  \institution{Curtin University}
  \city{Perth}
  \country{Australia}}
\email{shreya.ghosh@curtin.edu.au}

\author{Munawar Hayat}
\orcid{0000-0002-2706-5985}
\affiliation{%
  \institution{Qualcomm}
  \city{San Diego}
  \country{United States}}
\email{hayat@qti.qualcomm.com}

\author{Dimitrios Kollias}
\orcid{0000-0002-8188-3751}
\affiliation{%
  \institution{Queen Mary University of London}
  \city{London}
  \country{United Kingdom}}
\email{d.kollias@qmul.ac.uk}

\author{Kalin Stefanov}
\orcid{0000-0002-0861-8660}
\affiliation{%
  \institution{Monash University}
  \city{Melbourne}
  \country{Australia}}
\email{kalin.stefanov@monash.edu}

\author{Usman Tariq}
\orcid{0000-0002-8244-2165}
\affiliation{%
  \institution{American University of Sharjah}
  \city{Sharjah}
  \country{United Arab Emirates}}
\email{utariq@aus.edu}

\renewcommand{\shortauthors}{Zhixi Cai et al.}


\begin{abstract}
The detection and localization of deepfake content, particularly when small fake segments are seamlessly mixed with real videos, remains a significant challenge in the field of digital media security. Based on the recently released AV-Deepfake1M dataset, which contains more than 1 million manipulated videos across more than 2,000 subjects, we introduce the 1M-Deepfakes Detection Challenge. This challenge is designed to engage the research community in developing advanced methods for detecting and localizing deepfake manipulations within the large-scale high-realistic audio-visual dataset. The participants can access the AV-Deepfake1M dataset and are required to submit their inference results for evaluation across the metrics for detection or localization tasks. The methodologies developed through the challenge will contribute to the development of next-generation deepfake detection and localization systems. Evaluation scripts, baseline models, and accompanying code will be available on \url{https://github.com/ControlNet/AV-Deepfake1M}.
\end{abstract}





\maketitle

\def\thefootnote{}\footnotetext{Authors order is based on their surnames.}

\section{Introduction}
Rapid advances in generative AI have significantly transformed the landscape of digital content creation, making the generation and manipulation of video~\cite{singerMakeAVideo2022, geLong2022, wuTuneAVideo2023} and audio~\cite{shenNaturalSpeech2023, jiangMegaTTS2023, jiangMegaTTS2023a} more accessible and efficient than ever before. Although these technologies have enabled many positive applications in various domains, they have also created deepfakes, highly realistic but artificially manipulated media that can misrepresent people without their consent~\cite{zhouFace2021, narayanDFPlatter2023}. Deepfakes pose serious ethical and security concerns, as they can change individuals saying or doing things they never actually did, leading to widespread misinformation, disinformation, and even malicious activities such as online harassment and fraud~\cite{schwartzYou2018, brandonThere2019, sampleWhat2020, thomasDeepfakes2020}. The effectiveness of deepfake detection and localization methods is highly dependent on the datasets. Although the past few years have seen an increase in publicly available datasets focused on visual-only~\cite{jiangDeeperForensics12020, liCelebDF2020, kwonKoDF2021}, audio-only~\cite{liuASVspoof2023, yiADD2022}, and audio-visual~\cite{khalidFakeAVCeleb2021} content manipulations, most of these datasets assume that the entire content is either real or fake. This presents a significant limitation as it overlooks the growing trend of embedding small, subtle manipulations within otherwise real content. Such clever manipulations can completely alter the meaning of content, as discussed in~\cite{caiYou2022}, yet they are not adequately addressed by existing benchmark datasets and challenges.


To overcome the gap, the AV-Deepfake1M dataset~\cite{caiAVDeepfake1M2023} was introduced, providing a large-scale benchmark of audio-visual content-based deepfake videos for the task of temporal deepfake localization. Based on this dataset, the 1M-Deepfakes Detection Challenge not only focuses on the binary classification of deepfake content, but also emphasizes the task of localization, which means identifying the specific timestamps within a video where manipulations have occurred. The challenge is planned to contribute to improve current detection methods and aims to run as an ongoing benchmarking for the next several years, continually introducing new challenges of deepfake technology to keep pace with its rapid evolution.

\section{Related Work}
\begin{figure}[t]
\centering
\includegraphics[width=0.85\linewidth]{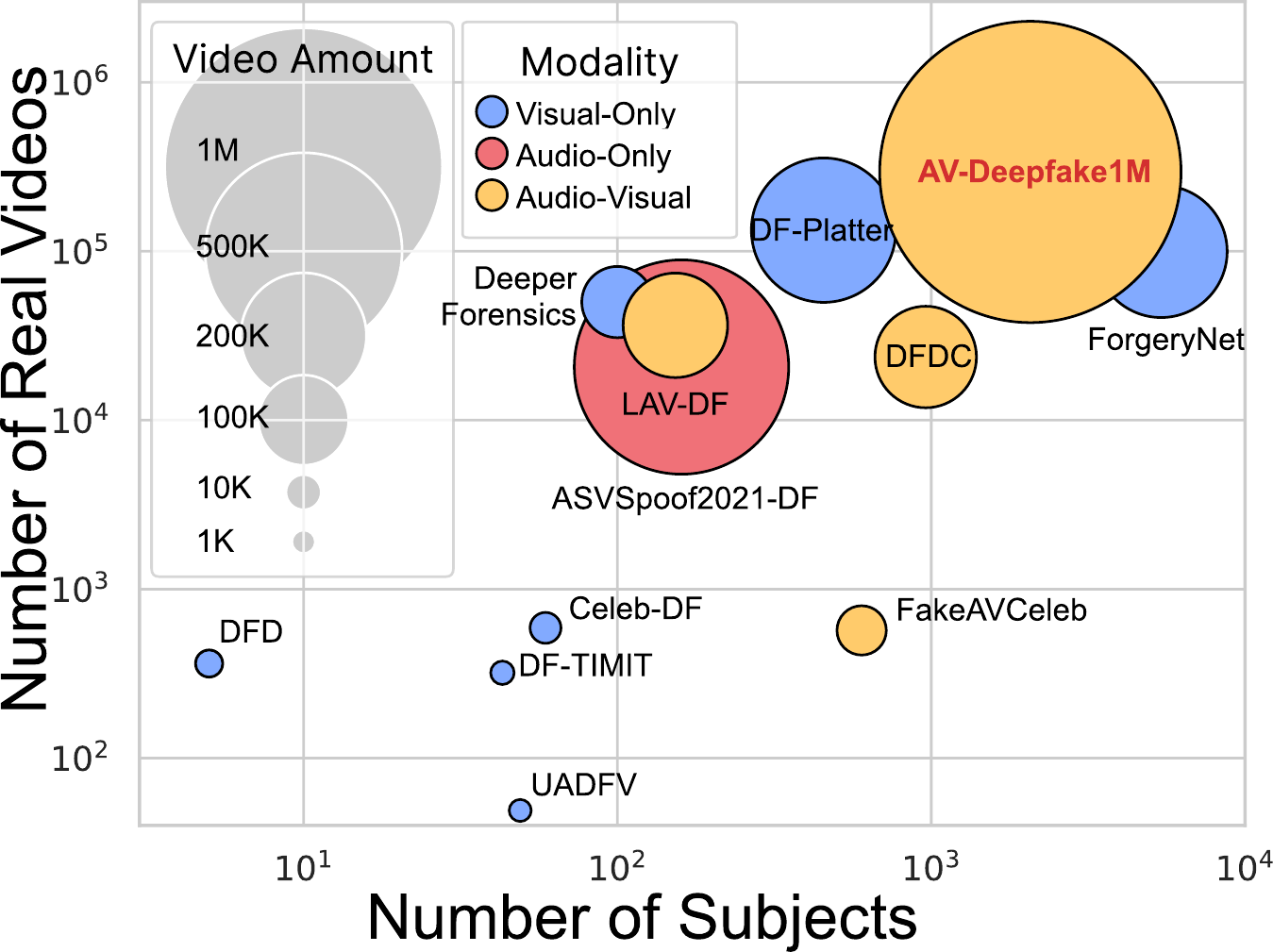}\vspace{-3mm}
\caption{\textbf{Comparison of related datasets with \datasetabbr{}.} \textmd{This figure illustrates a comparison of \datasetabbr{} with other accessible datasets, highlighting the number of subjects and the quantity of real versus fake videos. The figure is reproduced from the AV-Deepfake1M paper.}}
\label{fig:datasets}
\vspace{-5mm}
\end{figure}

Research on deepfake detection has progressed significantly due to the development of key datasets created using various manipulation techniques.
Figure~\ref{fig:datasets} outlines these relevant datasets.
Korshunov and Marcel~\cite{korshunovDeepFakes2018} were among the first to curate a deepfake dataset, DF-TIMIT, which involved face swapping on the Vid-Timit dataset~\cite{sandersonVidTIMIT2002}.
Subsequently, other notable datasets such as UADFV~\cite{yangExploring2018}, FaceForensics++~\cite{rosslerFaceForensics2019}, and Google DFD~\cite{nickContributing2019} were developed.
These datasets are relatively small due to the complexity of face manipulation and the limited availability of open-source manipulation techniques~\cite{liCelebDF2020}.

In 2020, Facebook introduced the large-scale DFDC dataset~\cite{dolhanskyDeepFake2020} for deepfake classification, comprising 128,154 videos created using various manipulation methods and featuring real videos of 3,000 actors.
DFDC has since become a standard benchmark for deepfake detection.
Following DFDC, several new datasets were released, including Celeb-DF~\cite{liCelebDF2020}, DeeperForensics~\cite{jiangDeeperForensics12020}, and WildDeepFake~\cite{ziWildDeepfake2020}, which focus on the binary task of deepfake classification and primarily target visual manipulation detection~\cite{chughNot2020}. 

In 2021, the OpenForensics dataset~\cite{leOpenForensics2021} was introduced for spatial detection, segmentation, and classification.
More recently, FakeAVCeleb~\cite{khalidFakeAVCeleb2021} was released, addressing both face-swapping and face-reenactment methods with manipulated audio and visual modalities.
ForgeryNet~\cite{heForgeryNet2021} is one of the latest additions to deepfake detection datasets, focusing on visual-only identity manipulations and supporting video/image classification as well as spatial and temporal forgery localization tasks.

In 2022, LAV-DF~\cite{caiYou2022} was introduced as the first content-driven deepfake dataset specifically designed for temporal localization.
However, LAV-DF suffers from limitations in both quality and scale.
Meanwhile, state-of-the-art methods for temporal localization~\cite{caiGlitch2023, zhangUMMAFormer2023} are already demonstrating very strong performance.
AV-Deepfake1M~\cite{caiAVDeepfake1M2023} addresses these limitations by enhancing the quality, diversity, and scale of the datasets available for temporal deepfake localization.

\begin{figure}[t]
\centering
\includegraphics[width=0.85\linewidth]{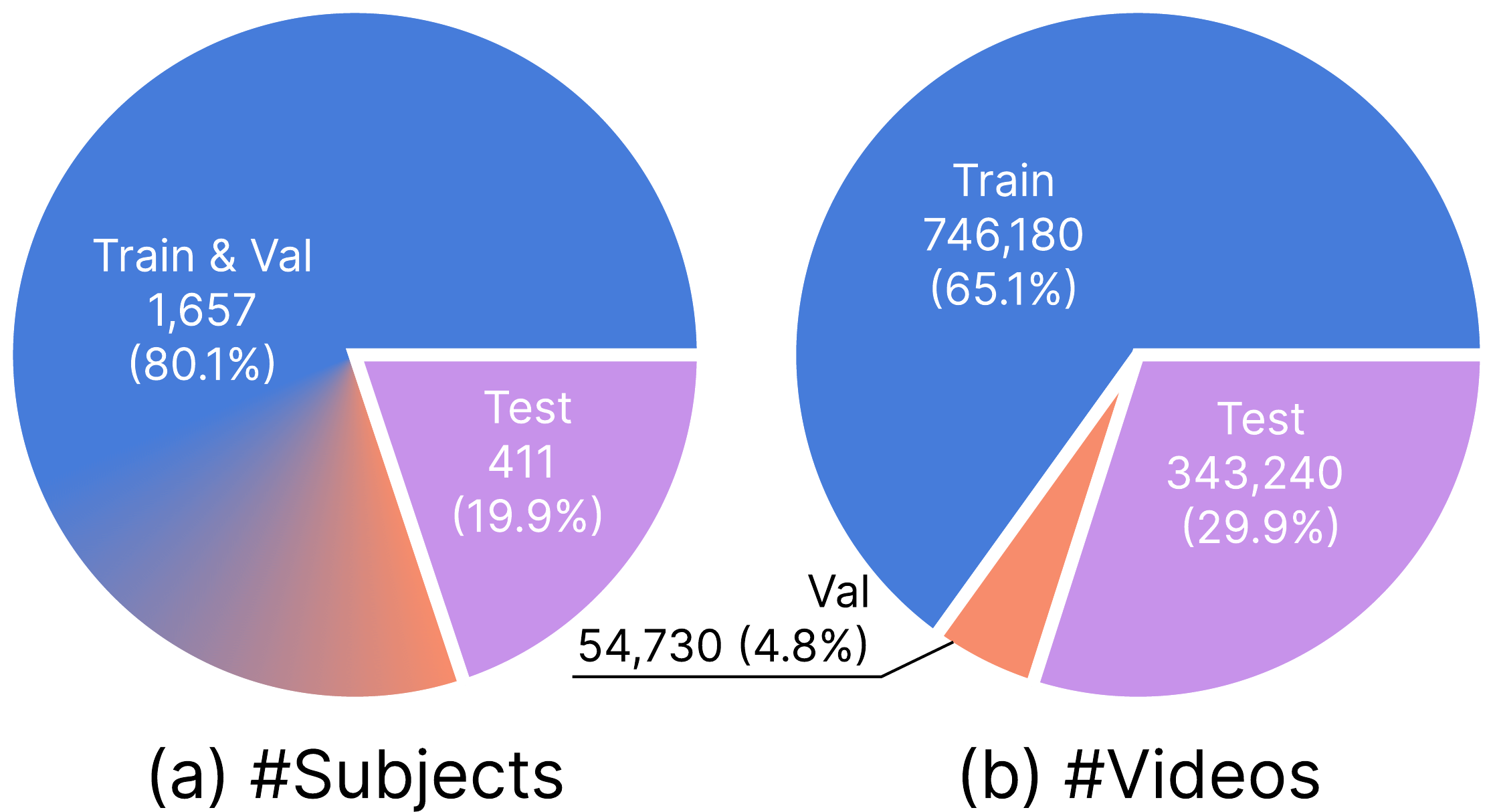}\vspace{-2mm}
\caption{\textbf{Data partitioning in \datasetabbr{}.} \textmd{(a) The count of subjects within the \emph{train}, \emph{validation}, and \emph{test} sets. (b) The count of videos present in the \emph{train}, \emph{validation}, and \emph{test} sets. The figure is adapted from the AV-Deepfake1M paper.}}
\label{fig:dataset_stats}
\vspace{-5mm}
\end{figure}

\section{Challenge Description}
\subsection{Dataset}
The challenge is based on \datasetabbr{} dataset~\cite{caiAVDeepfake1M2023}. The \datasetabbr{} dataset is a large-scale audio-visual deepfake dataset, containing 1,886 hours of audio-visual data from 2,068 unique subjects with diverse background settings.
This dataset is positioned as the most extensive audio-visual benchmarking effort, as shown in Figure~\ref{fig:datasets}.
The 1M-Deepfakes Detection Challenge involves two main tasks:

\noindent \textbf{Deepfake Detection.} This task requires identifying whether a given audio-visual sample of a single subject is deepfake or real.
\noindent \textbf{Deepfake Temporal Localization.} This task involves determining the specific time intervals within an audio-visual sample of a single subject that is manipulated or fake.

\begin{figure*}[t]
\centering
\includegraphics[width=0.85\textwidth]{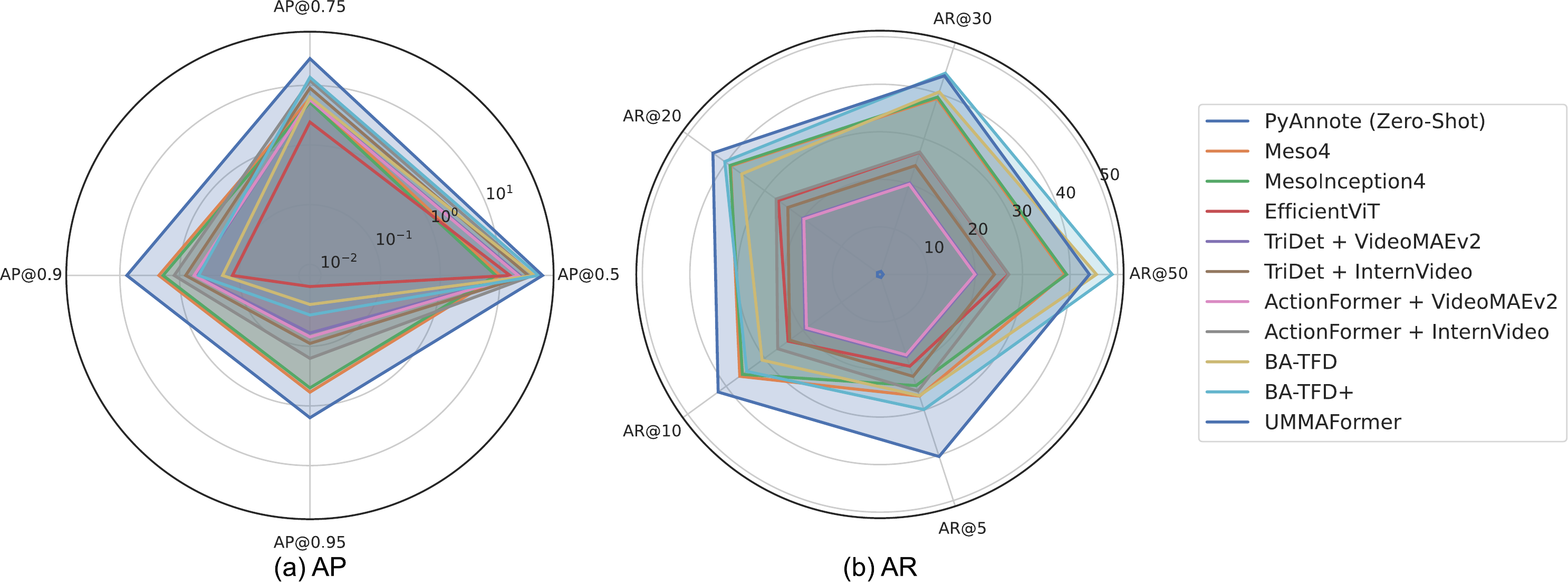}\vspace{-2mm}
\caption{\textbf{Temporal deepfake localization benchmark.} \textmd{This figure compares the performance of state-of-the-art methods on the \datasetabbr{} dataset.}}
\label{fig:temporal_localization}
\vspace{-3mm}
\end{figure*}

\begin{figure*}[t]
\centering
\includegraphics[width=\textwidth]{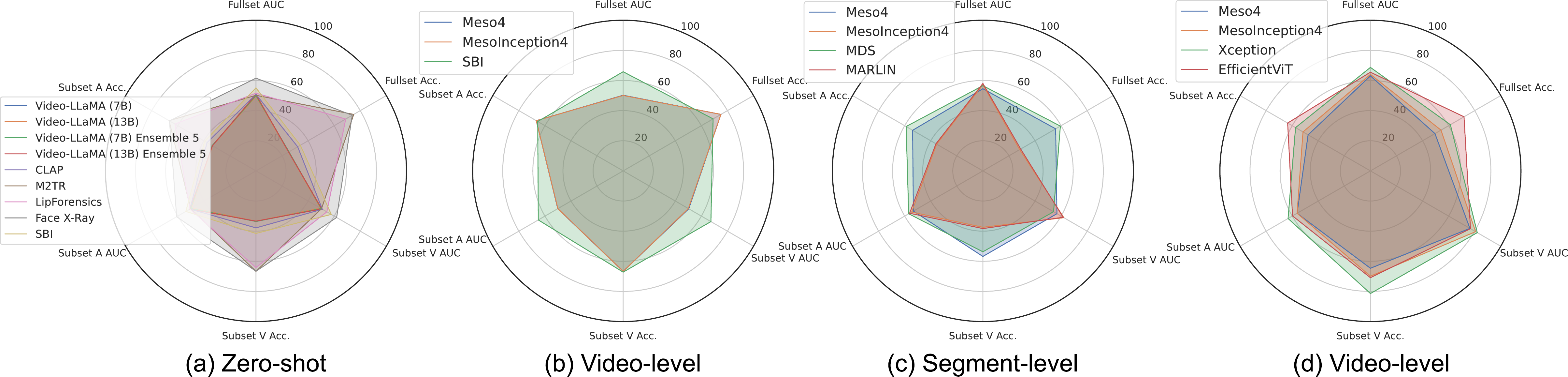}\vspace{-2mm}
\caption{\textbf{Deepfake Detection Benchmark.} \textmd{Comparison of state-of-the-art method performance on the \datasetabbr{} dataset across various evaluation protocols.}}
\label{fig:classification}
\vspace{-3mm}
\end{figure*}

\subsection{Data Partitioning}
To ensure a robust benchmark for the submitted models, we divided our dataset into train, validation, and test sets.
We randomly allocated 1,657 subjects to the train set and 411 subjects to the test set, ensuring no overlap between these groups.
The validation set was randomly selected from the train set based on the video level rather than the identity level.
For the test set, we ensured that all audio manipulations were generated using high-quality VITS models.
Finally, the distribution of the train, validation, and test sets is 65\%, 5\%, and 30\%, respectively (refer to Figure~\ref{fig:dataset_stats} for more details).

\subsection{Evaluation Metrics}
\subsubsection{Deepfake Detection} 
Deepfake detection is calculated as video label prediction.
For deepfake detection/classification, we use the area under the ROC curve (AUC).
The value is in the range of $[0.5, 1]$ where 0.5 denotes random guesses and 1 perfect prediction.

\subsubsection{Deepfake Temporal Localization} For deepfake temporal localization, we use average precision (AP) and average recall (AR) in line with previous works~\cite{caiYou2022, cabaheilbronActivityNet2015}. We set the intersection over union (IoU) thresholds at 0.5, 0.75, 0.9, and 0.95 for AP, and [0.5:0.95:0.05] for AR. For the number of proposals for AR, we select 50, 30, 20, 10, and 5. Finally, we average all AP metrics and all AR metrics to compute the overall average values for AP and AR. The average values of these metrics are then used to obtain the final results. The score ranges from 0 to 1, where 0 indicates random guesses and 1 represents perfect predictions. The final score $S$ is calculated as:

$$S = \frac{\sum_{IoU\in\{0.5,0.75,0.9,0.95\}}AP@IoU}{8} + \frac{\sum_{N\in\{50,30,20,10,5\}}AR@N}{10}$$


\subsection{Benchmark}
The results of the off-the-shelf models for deepfake temporal localization~\cite{plaquetPowerset2023, afcharMesoNet2018, coccominiCombining2022, shiTriDet2023, zhangActionFormer2022, wangVideoMAE2023, wangInternVideo2022, caiYou2022, caiGlitch2023, zhangUMMAFormer2023} and deepfake detection~\cite{zhangVideoLLaMA2023, wuLargeScale2023, wangM2TR2022, haliassosLips2021, liFace2020, shioharaDetecting2022, afcharMesoNet2018, chughNot2020, caiMARLIN2023, cholletXception2017, coccominiCombining2022} on the \datasetabbr{}~\cite{caiAVDeepfake1M2023} dataset are depicted in Figure~\ref{fig:temporal_localization} and Figure~\ref{fig:classification}.

\section{Challenge Participation Details}
In the 1M-Deepfakes Detection Challenge 2024, 191 teams signed the EULA and registered for the challenge.
The evaluation was carried out on CodaBench~\cite{xu2022codabench}, facilitating automated participation, submission, and evaluation.
By the submission deadline, there were 1034 successful submissions, indicating significant community interest. The number of submissions for Task 1 (detection/classification) and Task 2 (temporal localization) are not balanced.
Specifically, the number of submissions to Task 1 is 51\% higher than those to Task 2.
This imbalance suggests that the community focus is mainly on deepfake detection, thereby overlooking the significant threat posed by deepfakes that are not fully fabricated.
The latest work on deepfake detection methodologies further supports this trend~\cite{shaoDetecting2024,  oorloffAVFF2024, yangAVoiDDF2023, yuPVASSMDD2023, ilyasAVFakeNet2023, fengSelfSupervised2023, razaMultimodaltrace2023}.

\subsection{Challenge Track 1 (Detection/Classification)} In this track, the top 3 teams are Fake-detector (USTC), UQCV (UQ) and FRIdatas (FRI).
The Fake-detector (USTC) team mainly introduced the Audio-Visual Local-Global Interaction Module (AV-LG Module), which consists of Local Intra-Region Self-Attention, Global Inter-Region Self-Attention, and Local-Global Interaction, enhancing the fine-grained perception and cross-modal interaction capabilities of the model on deepfakes detection.
The key method for UQCV (UQ) team is weakly supervised learning via pseudo-labeling.
Using audio-based and mouth region-based pseudo-labeling, the proposed method gradually filters out high-confidence groups and adds those pseudo-labeled samples to the training set to fine-tune the network until all predictions are high-confidence.
Finally, the third-ranked team, FRIdatas (FRI) has developed an audio-visual method to detect fake chunks in the video.
A frame-based approach with a vision transformer model has been incorporated in a manner that mitigates the impact of different encoder versions in real and fake videos.
For audio, Wav2Vec-XLS-R features with an embedding module have been incorporated with temporal convolution operations.

\subsection{Challenge Track 2 (Temporal Localization)}
In this track, the top 3 teams are Gradiant (Gradiant)~\cite{vigo_deep}, Maya (USTC) and nudt24 (NUDT).
The top team, Gradiant, employs the gMLP method for audio features and complements it with a pathway that processes optical flow and RGB features using a UMMAFormer architecture.
The outputs from both pathways are merged to boost performance.
The Maya (USTC) team utilizes the BYOL-A model to extract audio features and both TSN and InternVideo models to extract visual features.
The BYOL-A audio features and the InternVideo visual features were used for model training in the video-level deepfake detection task, while the BYOL-A audio features and TSN visual features were used for the deepfake temporal localization task.
Finally, the third-ranked team nudt24 (NUDT) proposes a Task Hierarchical Emotion-aware for Fake Detection (THE-FD) architecture that initially handles video-level data (Task 1) and subsequently naturally adapts to frame-level data (Task 2).
The benefit of this approach is it allows for inheritance and sharing of features while utilizing a shared structure.

\section{Research Impact}
Existing deepfakes detection datasets (DFDC~\cite{dolhanskyDeepFake2020}, DF-TIMIT~\cite{korshunovDeepFakes2018}, Celeb-DF~\cite{liCelebDF2020}, etc.) have made a seminal contribution to the progress in deepfakes detection.
The current set of datasets is limited by their size and the range of tasks.
The latter refers to their focus on the binary classification task of deepfakes detection only.
For progressing the research in deepfake detection it is important to learn and evaluate the detectors on varied data representing a large number of subjects, situations, and contexts.
A comparison of the \datasetabbr{} with the existing datasets is presented in Figure~\ref{fig:datasets} (for a detailed comparison, please refer to \datasetabbr{}~\cite{caiAVDeepfake1M2023}).
The assumption that manipulation constitutes the entirety of a video has been shown to fail based on deepfake videos on social media, where a segment of the video is deepfaked.
The 1M-Deepfakes Detection Challenge will accelerate research in deepfakes analysis through the availability of both detection and localization tasks.

\section{Conclusion and Future Directions}
We introduced the 1M-Deepfakes Detection Challenge 2024, the first challenge addressing content-driven deepfake detection and localization in well-defined conditions.
We presented publicly available datasets and evaluation protocols for both tasks and evaluated baseline approaches.
The evaluation server will remain accessible to researchers beyond the 1M-Deepfakes Detection Challenge 2024 deadline, contributing to continuing progress on both tasks.

The task of identifying deepfakes goes beyond the typical classification approach. In this work, we explore both classification and localization, however, there are important aspects of deepfake analysis for future work. A major challenge in deepfakes comes from the variability due to different \textbf{cultures and languages}. From an observer's perspective, detecting deepfakes in non-native languages is non-trivial \cite{khan2023exploring}. Typically, foreign language videos are accessed and understood through voice-overs and native language subtitles. This makes the task of deepfake detection more challenging, as the semantic context and nuances of a particular culture or language may be missed by an observer. For a deepfake detector, scenarios like this require datasets containing content in different languages and methods that are agnostic to languages.

\textbf{Beyond Faces: } The current literature mainly focuses on the facial region in a video, as most deepfakes involve facial manipulation. The perceived meaning of a video can be altered for an observer with a simple edit to the non-verbal gestures of the subject's face and body. There is a need to explore this line of work. An example is the detection of sign language-based deepfakes \cite{naeem2024generation}. On the other hand, in scenarios involving groups of people, changing the group structure and/or the location of group members can lead to misinterpretation. Understanding group structures and the relationships between group members will be necessary to detect deepfakes in such conditions.



\bibliographystyle{ACM-Reference-Format}
\bibliography{main_cr}

\end{document}